\pdfoutput=1
\documentclass[a4paper]{svproc}
\usepackage{url}

\usepackage{float} 
\usepackage{tikz}
\usepackage{hyperref}
\usepackage{listings}
\usepackage{graphicx}
\usepackage{colortbl}
\usepackage{wrapfig}
\usepackage{footnote}
\makesavenoteenv{tabular}
\makesavenoteenv{table}
\usepackage{changepage}
\pdfoutput=1
\newcolumntype{L}{>{\centering\arraybackslash}m{3cm}}
\definecolor{champagne}{rgb}{0.97, 0.91, 0.81}
\usetikzlibrary{3d,decorations.text,shapes.arrows,positioning,fit,backgrounds}
\tikzset{pics/fake box/.style args={
#1 with dimensions #2 and #3 and #4}{
code={
\draw[gray,ultra thin,fill=#1]  (0,0,0) coordinate(-front-bottom-left) to
++ (0,#3,0) coordinate(-front-top-right) --++
(#2/2,0,0) coordinate(-front-top-right) --++ (0,-#3,0) 
coordinate(-front-bottom-right) -- cycle;
\draw[gray,ultra thin,fill=#1] (0,#3,0)  --++ 
 (0,0,#2/4) coordinate(-back-top-left) --++ (#2/2,0,0) 
 coordinate(-back-top-right) --++ (0,0,-#2/4)  -- cycle;
\draw[gray,ultra thin,fill=#1!80!black] (#2/2,0,0) --++ (0,0,#2/4) coordinate(-back-bottom-right)
--++ (0,#3,0) --++ (0,0,-#2/4) -- cycle;
\path[gray,decorate,decoration={text effects along path,text={CONV}}] (#2/4,{2+(#3-2)/2},0) -- (#2/6,0,0);
}
}}
\tikzset{circle dotted/.style={dash pattern=on .05mm off 2mm,
                                         line cap=round}}

\begin{document}
\mainmatter              
\title{Embedded Neural Networks for Robot Autonomy}
\titlerunning{Embedded Neural Networks for Robot Autonomy}
%
\author{Sarah Aguasvivas Manzano\inst{1} \and Dana T. Hughes\inst{2} \and Cooper R. Simpson\inst{1} \and Radhen Patel \inst{1} \and Nikolaus Correll\inst{1}}
\authorrunning{Aguasvivas Manzano, S., Hughes, D., Simpson, C., Patel, R., Correll, N.} 
%
\tocauthor{Aguasvivas Manzano, S., Hughes, D., Simpson, Patel R., C., Correll, N. }
\institute{University of Colorado Boulder, Boulder CO 80309, USA
\and
Carnegie Mellon University, Pittsburgh, PA, 152013, USA}

\maketitle 

\begin{abstract}
 We present a library to automatically embed signal processing and neural network predictions into the material robots are made of. Deep and shallow neural network models are first trained offline using state-of-the-art machine learning tools and then transferred onto general purpose microcontrollers that are co-located with a robot's sensors and actuators. We validate this approach using multiple examples: a smart robotic tire for terrain classification, a robotic finger sensor for load classification and a smart composite capable of regressing impact source localization. In each example, sensing and computation are embedded inside the material, creating artifacts that serve as stand-in replacement for otherwise inert conventional parts. The open source software library takes as inputs trained model files from higher level learning software, such as Tensorflow/Keras \cite{tensorflow2015,chollet2015keras}, and outputs code that is readable in a microcontroller that supports C. We compare the performance of this approach for various embedded platforms. In particular, we show that low-cost off-the-shelf microcontrollers can match the accuracy of a desktop computer, while being fast enough for real-time applications at different neural network configurations. We provide means to estimate the maximum number of parameters that the hardware will support based on the microcontroller's specifications.

\end{abstract}

\section{Introduction}

Ongoing miniaturization of computation and advanced manufacturing technologies will blend the distinctions between a robot's ``body'' and a robot's ``brain'' \cite{Menguc,Correll_2017}. At the same time, advances in machine learning have facilitated the implementation of complex signal processing and control by learning from examples and/or rewards \cite{goodfellow}. As a consequence, machine learning has supplemented and sometimes replaced model-based identification and control techniques to capture the non-linear relationships between inputs and outputs of robot sensors, mechanisms and actuators.


    

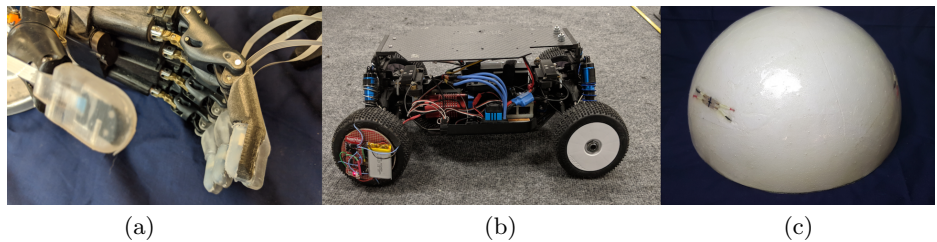
\begin{figure}[H]
\centering
 \includegraphics[width=\textwidth]{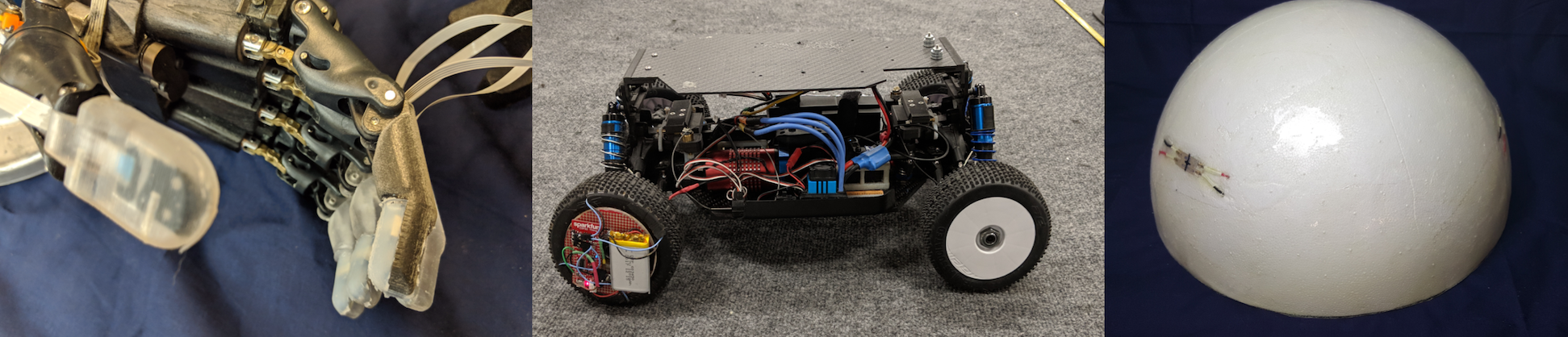}
  \begin{picture}(100, 10)
    \put(-80, 0) {(a)}
    \put(55, 0) {(b)}
    \put(167, 0) {(c)}
  \end{picture}
  \caption{Robotic materials that showcase the feasibility of embedding intelligence closer to the material where data is being collected from: (a) prosthetic hand with tactile sensors on the fingertips, (b) self-driving car with a terrain sensitive tire, (c) composite with embedded sensors for impact source localization.}
  \label{fig:applicationfig}
\end{figure}

While our understanding on how to train models for a large number of prediction and controls problems continues to increase at a fast pace, the prevailing tools tend to reinforce the sense-plan-act paradigm in which sensing information is gathered in a central location and controls are computed and relayed back to actuators in the system. Instead, we envision processing and control to be deeply embedded in the (composite) material itself \cite{mcevoy2015materials}, an abstraction which we hope to dramatically facilitate the construction of autonomous robots \cite{Correll_2017}. Under this abstraction, every part of the robot is intelligent and capable of performing complex computations that supplement the central processing of the robot. In addition to providing a powerful design abstraction, this approach might also be the only one to deal with hard constraints that the robot's embodiment imposes. In addition to the geometry and dynamics of the robot's structure and sensor placement \cite{pfeifer2007self}, central processing of information comes with additional challenges in routing information, introducing latency, bandwidth limitations, overhead, and manufacturing challenges, thereby imposing artificial constraints on signal processing and control. 
The human body, for example, addresses these latency problem by offloading some computation from the central nervous system as seen in our enteric nervous system \cite{hadhazy2010think}.

Transferring signal processing and control algorithms into material-embedded hardware platforms addresses several of the above issues; however, implementing the algorithms on embedded, general purpose microcontrollers remains a hard software engineering challenge, often requiring specialized experience for each target platform.  Deep neural networks highlight this challenge:  machine learning experts design, implement, and train deep networks using high level languages (e.g., Python), rich frameworks (e.g., Tensorflow), and abstract datatypes (e.g., tensors), while embedded platforms utilize low-level languages (e.g., C and assembly) and simple memory management (e.g., fixed-sized arrays).  This mismatch makes representing and implementing deep neural networks in microcontrollers a very laborious task.  This prevents quick development cycles that involve holistic dynamics and environment interactions, which are critical in the robotic design process.




Once a model is trained, the learned parameters need to be copied into code that is adapted to the microcontroller's architecture; the neural network architecture must match exactly the original model for our estimations to have fidelity with the model that was trained. Any mistake in this process might result in incorrect estimations, memory leaks or additional debugging time. We believe that it is challenges of these kinds that must be circumvented urgently in order to tackle the tight integration of sensing, actuation, computation and material physics that natural systems use to enable autonomy.

\begin{figure}[H]
\centering
\includegraphics[width=\linewidth]{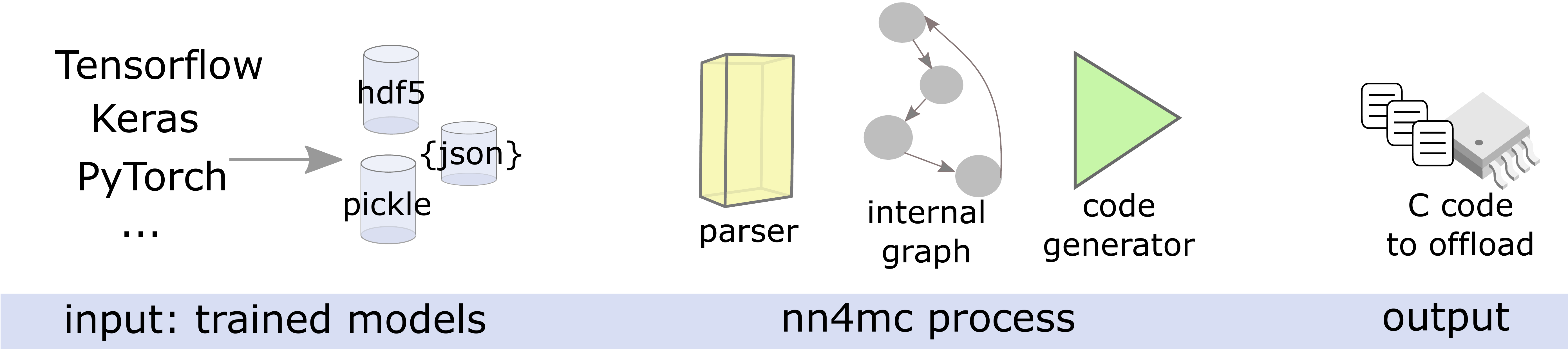}
  \caption{\textbf{\textit{nn4mc} approach}. There are three major steps in this approach: (1) collecting data and loading into a PC, (2) training neural network and compiling the feed forward code in \textit{nn4mc}, (3) off-loading code into target microcontroller.}
  \label{fig:pipeline}
\end{figure}

To this end, we present the tool \textit{nn4mc}, which is available open source\footnote{\url{https://github.com/correlllab/nn4mc}}, to transfer deep learning models onto small, general purpose microcontrollers.  These microcontrollers can then be co-located with sensors and actuators distributed throughout a composite material or robot.  Here, we repurpose off-the-shelf microcontrollers that are already being used to collect and preprocess sensor signals or implementing low-level feedback control.

\subsubsection{Overview of the proposed approach}
Figure \ref{fig:pipeline} illustrates the usage process proposed in this work. After recording data from physical hardware, the training of the neural network is performed in a PC with high level language capabilities (e.g. Keras \cite{chollet2015keras}, Tensorflow \cite{tensorflow2015}, PyTorch \cite{paszke2017automatic}), that may have GPU features to accelerate a computationally intense training process. The resulting networks, including the learned weights and biases, can then be parsed into \textit{nn4mc}, which we describe in Section \ref{sec:parse}. We then proceed to generate the C code using \textit{nn4mc} that can be integrated as a simple function, not requiring any additional libraries to represent the code or write the code.  
In Section \ref{sec:perf} we break down some performance details of this function implemented for different neural network architectures and validate that the output of the embedded solution is on par with that trained on the desktop computer. Section \ref{sec:mpc} is an exploration on how \textit{nn4mc} can be introduced in a Model Predictive Control (MPC) problem by making a neural network learn the system identification.

\subsubsection{Related Work} In the past, researchers have explored both software-based and hardware-based approaches for embedded intelligence. Regarding software-based approaches, the concept of Edge IoT \cite{IoT-principles} has pushed towards shifting computation on-board as opposed to having an all-mighty cloud receiving and processing data gathered by a microcontroller. Other software-based online prediction models look into different ways to obtain online machine learning predictions on microcontrollers. Regarding hardware-based approaches, we find another handful of work done  \cite{hardware-ANN,cotton_2010,uTensor,tensorflowLite}. The former two are software-based approaches, but have a limited range of target microcontrollers and propose a dedicated microcontroller for optimized operations. We found many disadvantages on the existing methods. To name a few we have: (1) some of the packages are not open source or currently available \cite{fraunhofer_ims}, so they create executable files instead of code that will not allow for a full integration with existing code \cite{hardware-ANN,chem_e_nn} (2) Some of them do not try to repurpose existing microcontrollers, instead they suggest a dedicated microcontroller or IDE that can read the code \cite{hardware-ANN,cotton_2010,uTensor,tensorflowLite}. (3) There are cases in which training within the microcontroller is not the goal \cite{hardware-ANN,neural_online}, as seen in the systems in Section \ref{sec:validation}, so we can rely on higher computation power coming from PCs for training the neural network. (4) Other cases were designed for a one time use and require major coding to adapt to different architectures \cite{ANN_mpc}.

The Wireless Sensor Network (WSN) community has started to shift towards more self-sufficient microcontrollers \cite{Self_sufficient} capable of performing online neural network predictions. Currently, the most common approach towards this goal is to create dedicated hardware that is specialized on a few specific neural network architectures \cite{hardware}. The proposed approach differs with the existing ones in the fact that it repurposes existing systems and microcontrollers to integrate the intelligence without the need to purchase a specific microcontroller or the need to do all the training within the microcontroller. 

\subsection{Problem Statement}
We would like to: (1) utilize standard, state-of-the-art frameworks to develop and train neural network models, and translate the learned models to specific embedded platforms; (2) develop an approach that is aware of memory constraints and processing capabilities, and select an appropriate representation (e.g., 8-bit ints, 32-bit floats) based on available hardware; (3) minimize as much as possible the computing time required to perform a forward pass of the network on the embedded architecture, to ensure that real-time deep neural network control is plausible; (4) repurpose existing hardware to embed the intelligence in it. 

\subsection{Contribution of this paper}
\textit{nn4mc} responds to these needs. In particular, \textit{nn4mc} is capable of producing code usable in \emph{multiple platforms}, as long as the platform can be written in C code. in Section \ref{sec:perf} we showcase multiple, commonly used, general purpose microcontrollers and their performance using this software. At the same time, \textit{nn4mc} is capable of processing outputs from many \emph{different sources}, who adhere to standardized file formats to represent neural networks. If a user desires to add a new feature to \textit{nn4mc}, all they need to do is create this feature as a derived class of the more general classes. The same holds true in the case that the user desires to add a new target, making our approach \emph{scalable}. Finally, \textit{nn4mc} is free and  \emph{open source} software. The proposed framework is validated using data from four robotic hardware demonstrators, and benchmarked on four different microcontroller architectures. 

\section{Software Architecture}
Our software architecture consists of three main modules: a \textit{Parser} module used to load neural network models and parameters from a variety of storage formats; an \textit{Abstract Representation} module used to represent and process neural network models; and a \textit{Code Generator} module that generates a set of source code files for implementing the neural network on a microcontroller.  We have based our architecture on common object-oriented design patterns~\cite{design_patterns}, which simplifies extending \textit{nn4mc} to include support for new deep learning framework formats, model and parameter file formats, and target microcontrollers.


To achieve the multiple-source-multiple-target goal, we convert the neural network model into an abstract graph representation of the neural network (\textit{Neural Network}) to then generate the code. This interface allows us to maintain our own format and have \textit{Parser} and \textit{Code Generator} adapt their inner workings to fit this abstract representation without having to interact with each other. 

\subsubsection{Data structures}
We create our own data structures  to maintain uniformity and control on the internal representation of data within \textit{nn4mc}, while mininimizing memory and computational requirements:

\begin{enumerate}
    \item \textit{Tensor} stores highly dimensional matrices and tensors in a single string; it also contains the necessary operations to map this string into usable vectors and matrices. 
    \item \textit{Weight} is a collection of a \textit{Tensor} pointer and a layer id. 
    \item \textit{Layer} is an abstract class that contains multiple derived classes, which are the different layer types that a neural network can have. This currently includes \textit{Conv1D}, \textit{Conv2D}, \textit{MaxPooling1D}, \textit{MaxPooling2D}, and \textit{Flatten}. 
    \item \textit{NeuralNetwork} is a directed graph that represents a trained model as connections between layers which act as the nodes. It also associates all data structures related to a layer with the corresponding node and it is the central data structure of the software. 
\end{enumerate}

\subsubsection{Parser}
\label{sec:parse}
\textit{Parser} is an abstract class created to load trained models from any given format into \textit{nn4mc}. We use \textbf{HDF5C++} \cite{hdf5} and \textbf{nlohmann/json} \cite{nlohmann_json} to parse \textit{.hdf5} and \textit{.json} files and transverse through the layer graph represented in binary in C++. The parser loads the neural network architecture along with the weights from a file exported from a neural network training package. It is the interface between the model and the abstract representation. This class breaks down the parsed model into the components needed to create and link the layer nodes that \textit{NeuralNetwork} needs. Its \textit{build\_nn()} method exports a pointer to a \textit{NeuralNetwork} object from a parsed object which is the abstract graph representation needed to generate the feed forward code. 
New support for  different neural network model file formats may be added through derived classes from \textit{Parser}. An example on how to use the parser class is under the example folder in the repository. Each parsing type requires two additional modules: \textit{LayerBuilder} and \textit{LayerFactory}. 



\subsubsection{LayerBuilder} This class links the parsed attributes of a specific layer with the attributes of the layer in our internal neural network representation. This class has derived classes, such as \textit{Conv2DBuilder} and \textit{DenseBuilder}, which individually create a pointer to a layer object and their associated attribute definitions.

\subsubsection{LayerFactory} This class contains an associative container (\textit{std::map}) that maps from the parsed layer type name to a pointer to a \textit{LayerBuilder} object. For instance, \textit{Parser} read the first layer and found out that this layer is a Dense layer; then it looks up the corresponding \textit{Layer} type that it needs to create, \textit{LayerFactory} allows \textit{Parser} to instantiate a \textit{LayerBuilder} based on that layer type that it parsed. 

\subsubsection{Abstract Representation}
The abstract representation of a trained model loaded through \textit{Parser} is achieved through the \textit{NeuralNetwork} class -- a graph data structure. This abstract representation allows for two important features of the \textit{nn4mc} software package. First, the code generation portion of this package needs only to interact with the abstract representation to obtain the necessary data for building the resultant code. This allows for the trained model to be imported in a multitude of formats. \textit{Parser} handles the reading of this input and building of the \textit{NerualNetwork} object which decouples the two halves of \textit{nn4mc} -- parsing and code generation. This makes \textit{nn4mc} flexible in its usage and scalable in scope. Second, the abstract representation of a trained neural network allows us to analyze its qualities and potentially manipulate its structure.

\subsubsection{LayerNodes}
\textit{LayerNode} is an essential substructure of the overall \textit{NerualNetwork} class. They represent the layers of the neural network and contain all associated data necessary for traversal and code generation. Most important in the layer representation is a pointer to a \textit{Layer} object created in the parsing that each \textit{LayerNode} structure contains. Similarly, each \textit{LayerNode} contains a list of the layers that are inputs into that specific layer. The \textit{Layer} object pointer allows one to extract weight and bias data. The list of inputs informs one of the necessary data needed before feeding through any particular layer. Both of these data are needed in order to properly generate the output code.

\subsubsection{Graph Traversal}
In order to use the Neural Network represented by a \textit{NeuralNetwork} object one must be able to push data through in the proper order. Using a Breadth First Traversal we iterate through the graph to understand the required flow of data for the code generation. A special kind of \textit{Layer} named \textit{Input} allows one to mark the entry point of a graph. From there, data follows the path of the directed edges. The graph is traversed in a Breadth First manner to ensure that that each layer has its required data before moving on in a forward feed. This allows the code generation to correctly order its processes.

\subsubsection{Code Generator}

Finally, the abstract representation of the neural network is translated into instructions for the microcontroller. For each target platform, we have a set of template files. \textit{Code Generator} formats the data stored in the \textit{Tensor} objects into weights and the layer data that each graph node is pointing at into neural network feed forward code.

\section{Results}
\subsubsection{Experiment Design}
We evaluate the neural network code generated by \textit{nn4mc} based on the \textit{computing time} required for the implemented network, and the relative \textit{accuracy} of the output of a model generated by \textit{nn4mc} with respect to the implementation in the original framework (Tensorflow).  We perform this evaluation on a set of example network architectures and embedded system platforms.


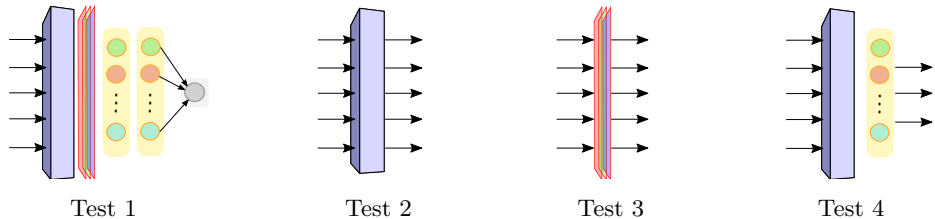
\begin{figure}[H]
      \centering
      \includegraphics[width=\textwidth]{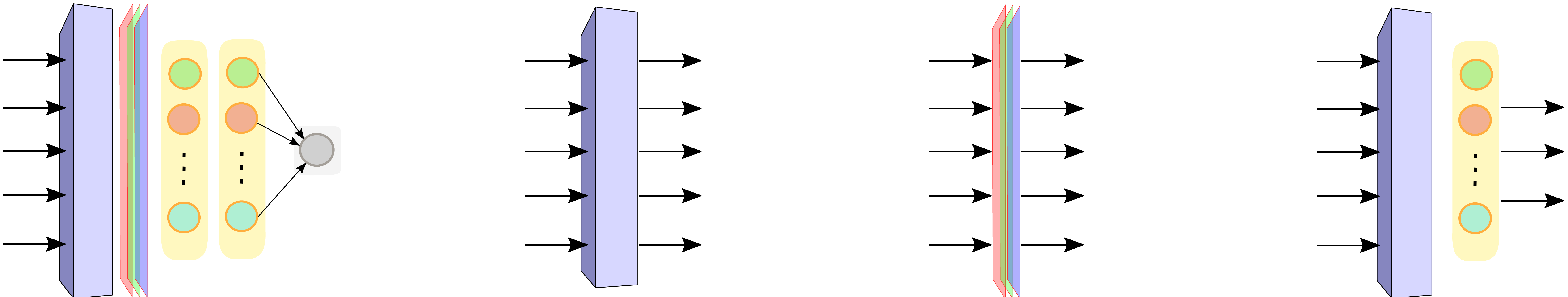}
      \begin{picture}(100,-10)
      \put(-100,-3){Test 1}
        \put(3,-3){Test 2}
          \put(90,-3){Test 3}
          \put(180,-3){Test 4}
       \end{picture}
      \caption{Graphical representation of the tests shown in Section \ref{sec:perf}. We used a purple box to represent 1-D convolution, the red filters to represent maxpooling and the yellow layers to represent fully connected layers.}
      \label{fig:test_fig}
\end{figure}

Figure \ref{fig:test_fig} illustrates the different neural network layer configurations that we test in this work. \emph{Test 1}, a CNN with one convolutional layer, a maxpooling layer, and three dense layers, is the neural network configuration used in the smart composite usage case seen in Section \ref{sec:validation} and a realistic usage case for deep neural networks in a microcontroller. \emph{Test 2}, a single convolution layer with 3 filters and a kernel size of 5 units. \emph{Test 3}, a single maxpooling layer with a pool size of 5 units, isolates a 1-D max-pooling layer, which is as complex of a mapping as convolution layer, but does not require to look at the registry for parameters. \emph{Test 4}, a small CNN with one convolution layer and one dense layer with an input size of 10 nodes, illustrates a simplified convolutional neural network that can be run in all the microcontrollers because it needs less parameters than Test 1.

\begin{table}
\begin{center}
\label{tab:summary}
\caption{Basic specifications for the development boards put to test in these experiment designs.}
\begin{tabular}{cccccccc}
\hline
\multicolumn{8}{c}{ \textbf{Summary of Development Boards}}  \\
\hline
\textbf{Case}& \textbf{Microcontroller} & \textbf{RAM}  & \textbf{Flash} &\textbf{EEPROM} & \textbf{Bits} & \textbf{Freq.} & \textbf{FPU}\\
\hline
A & ATmega328P\footnote{\href{https://store.arduino.cc/arduino-uno-rev3}{Arduino UNO}} & 2KB & 32 KB &  1 KB & 8 & 16 MHz & No\\
B & MKL26Z64VFT4\footnote{\href{https://www.pjrc.com/teensy/techspecs.html}{Teensy L-C}} & 8 KB & 62KB & 128 KB\footnote{emulated by software} & 32  & 50 MHz & No \\
C & MK66FX1M0VMD18\footnote{\href{https://www.pjrc.com/teensy/techspecs.html}{Teensy 3.6}} & 256 KB & 1024KB & 4KB & 32  & 180 MHz & Yes \\
D & Tensilica Xtensa LX6\footnote{\href{https://www.espressif.com/sites/default/files/documentation/esp32_datasheet_en.pdf}{ESP32}} & 520 KB  & 4MB & 448KB & 32 & 160 MHz & Yes\\
\hline
\end{tabular}
\end{center}
\end{table}

We tested the various network models on four different development boards, which are shown in Table \ref{tab:summary}. For this comparison, we focus on the attributed critical for neural network implementation, namely memory (RAM, EEPROM, and Flash), register width (bits), and clock frequencies. 


\subsection{Computing Time}
\label{sec:perf}
We first measure the computation time for all of the microcontrollers in each of the test cases using 1000 samples. We note that there is a single execution thread used in platforms A---C, while the embedded real-time operating system (RTOS) used in platform D schedules multiple threads, which resulted in an increased variance in the measured execution time due to possible context switches.

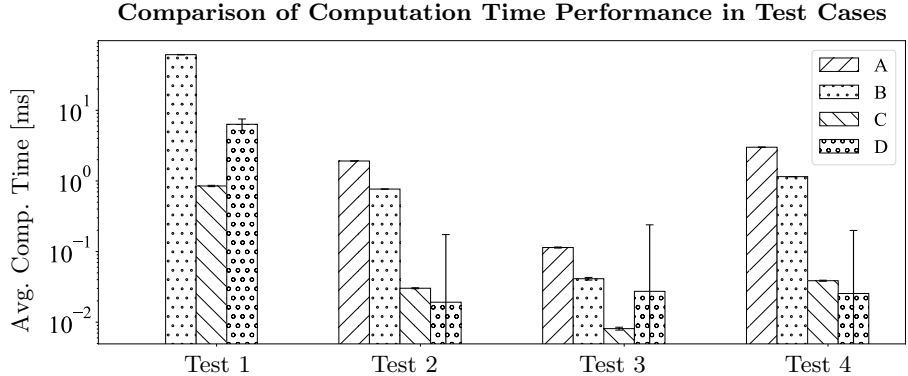
\begin{figure}
      \centering
      \includegraphics[width=\linewidth,trim={2.1cm 0.8cm 3.1cm 0.8cm},clip]{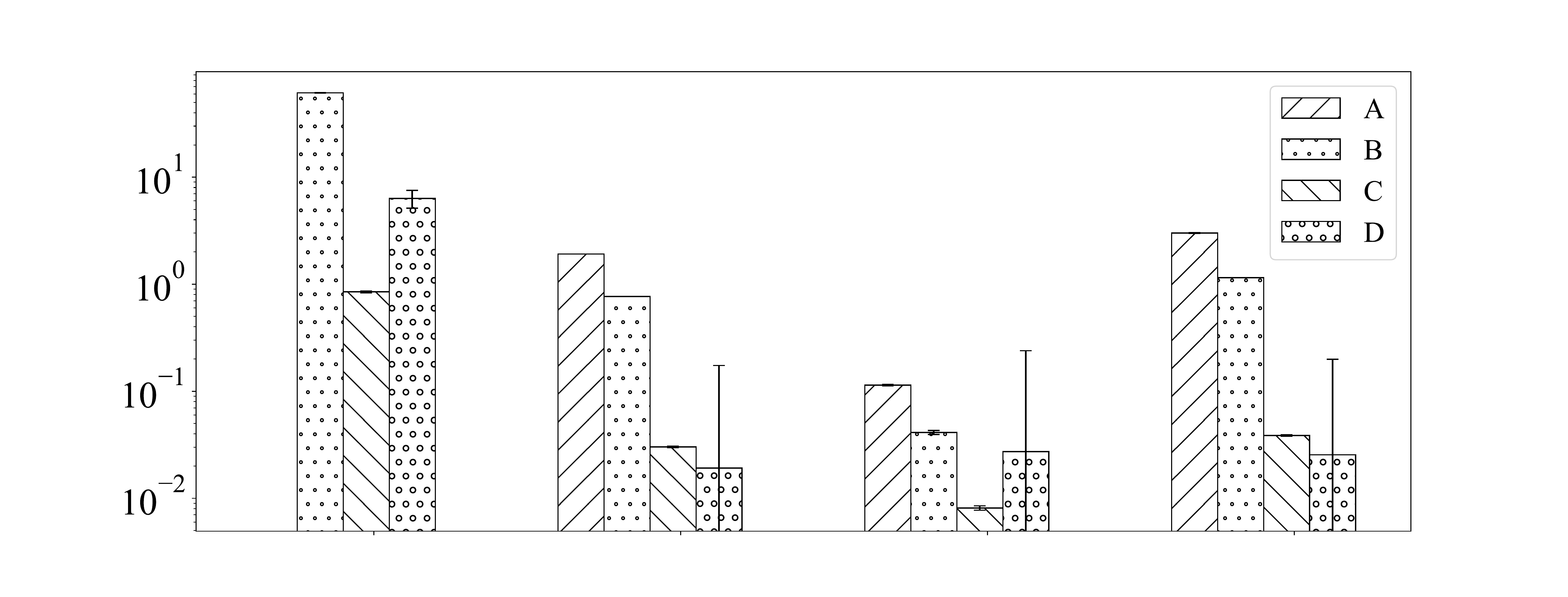}
      \begin{picture}(90,-15)
      \put(-65,10){Test 1}
        \put(5,10){Test 2}
          \put(83,10){Test 3}
          \put(160,10){Test 4}
          \put(-90,143){\textbf{Comparison of Computation Time Performance in Test Cases}}
          \put(-130, 25){\rotatebox{90}{Avg. Comp. Time [ms]}}
       \end{picture}
      \caption{Average computation times obtained throughout the tested neural network architectures and functions. The Arduino UNO (A) does not have enough memory to perform test 1 and has been omitted.  Note that computing time (y-axis) is in a log scale.}
      \label{fig:perf_fig}
\end{figure}

Figure \ref{fig:perf_fig} summarizes the performances of the microcontrollers tested in the four test cases performed in this work.  Results for platform A in test 1 have been omitted due to memory limitations of the Arduino UNO platform (see also Section \ref{sec:footprint}). The second least powerful platform (Teensy L-C, B) was able to perform the neural network prediction, but at about 61 ms, which might not be desirable for real time predictions, whereas the Teensy 3.6 (C) and ESP32 (D) perform well below $10ms$, making these platforms suitable for most control tasks. All platforms have a low variance; the highest variance is $0.18 ms$ (platform D) due to RTOS running a scheduler. 

\subsection{Fidelity at Fixed Point}
\label{sec:fixed}
A major limiting factor with using embedded microcontrollers for signal processing and machine learning is that few microcontrollers implement a floating point unit.  Instead, floating point operations are synthesized using the available instruction set (at a significant increase in computing time), or fixed point math is used.  

We investigate the case in which a microcontroller cannot operate at 32-bit floating point to forward pass data through a neural network. We compare based on the average of the difference between fixed point outputs for $k$ bits ($output_{k}$) and floating point outputs over all of the elements in the output of the neural network or layers we are trying to investigate.  $\epsilon_{k [bits]}$ represents the absolute errors reported in Table \ref{tab:fidelity}. This loss is calculated with respect to a fixed point representation of 32 bits. We find that at 16 bits, the average error is relatively low, considering that the maximum computed value in the output layer in this case is about 0.797.  

\begin{table}
\caption{Fidelity losses with respect to original model for different fixed point operations depending on the bits used.}
\label{tab:fidelity}
\setlength{\tabcolsep}{12pt}
\begin{center}
\begin{tabular}{cccc}
\hline
\multicolumn{4}{c}{\textbf{Fixed Point Errors ($\epsilon_{k}$)}}  \\
\hline
\textbf{Test}& \textbf{$\epsilon_2$} & \textbf{$\epsilon_8$} & \textbf{$\epsilon_{16}$}\\
\hline
1 & $0.04517$ & $0.00038$ & $<0.00001$ \\ 
2 & $0.11662$     & $0.00109$ & $<0.00001$ \\ 
3 & $0.26249$ &$0.00339$ &$<0.00001$ \\ 
4 & $1.07756$ & $0.01229$  & $0.00006$  \\ 
\hline
\end{tabular}
\end{center}
\end{table}

\subsection{Memory footprint}
\label{sec:footprint}

As the number of parameters in the neural network increases, memory requirements also increase. We estimate that if we are trying to use a fraction of the static memory $\gamma$, in a \textit{float32} representation, the maximum number of parameters $P$ that our neural network is allowed to take for our flash memory limitation $S$ in bits is given by $ P \leq \lfloor \frac{\gamma S}{32} \rfloor$.

For example, platform A has $S = 8,192$ $bits$ of memory (Table \ref{tab:summary}), with $\gamma=1$ our maximum number of parameters should be $P \leq 256 $. Indeed, Platform A is able to store the neural network of Test 4 (69 parameters), whereas the neural network in Test 1 ($\gg 256$ parameters) exceeds the available memory. 

\subsection{Validation of \textit{nn4mc} using robotic applications}
\label{sec:validation}

We demonstrate multiple instances in which \textit{nn4mc} has been useful to perform online neural network predictions. The neural networks shown in these examples have 1:1 fidelity between the PC models and the embedded version; therefore the online neural network performance on these systems is dependent on the data collection and handling and not on \textit{nn4mc}. We first demonstrate a terrain sensitive tire that classifies the terrain it is driving on. We then demonstrate regression on the localization of the source of an impact in a robotic skin and a smart composite. Finally, we show functioning of \textit{nn4mc} in a calibrated force given the raw sensor signals from a tactile sensor integrated into a prosthetic hand. Whereas these case studies have been also developed in previous work \cite{Dana_terrain,doi:10.1177/1687814019844643}, signal processing and transfer of neural networks so far has been either hand-coded, or not fully deployed in the data collection hardware. 

\subsubsection{Terrain Sensitive Tires}
\label{sec:tire}

In \cite{Dana_terrain} a classification neural network was embedded manually into a microcontroller in order to perform online predictions on the type of terrain that a smart tire is driving on, regardless of the driving modality of the pilot. In this work, we automatically generate a neural network-based classifier using \textit{nn4mc}. We then quantify the online neural network prediction performance into an \textit{Online Confusion Matrix}. In this work, we compare the testing set predictions (Table \ref{tab:conf_tire}) with the online predictions (Table \ref{tab:online_conf_tire}) to see if the testing set predictions actually guarantee a comparable performance in real time. Fig. \ref{fig:applicationfig}(b) shows the integration of the smart tire with a model all-terrain race car. 

In the training phase, the final accuracy for the testing set data was 83.90\%. In the online phase, the time it takes to make a single forward pass in this neural network, whose number of parameters amounts to $20,582$, is an average of $0.041 ms \pm 0.007 ms$ using the ESP32 platform (D). After the microcontroller on the tire is enabled to make online predictions, we isolate each of the scenarios that represent a particular class and store the neural network prediction results to collect data about the online accuracy of the prediction. For each forward pass, a data window of 400 samples (100 coming from each sensor) is fed into the neural network. 

 Table \ref{tab:conf_tire} represents the classification accuracy and precision in the testing set data which is equivalent to Table I in \cite{Dana_terrain}. Table \ref{tab:online_conf_tire} represents the classification accuracy of the online predictions.

\begin{table}
\begin{minipage}{0.55\linewidth}
\caption{PC results on collected dataset.}
\begin{center}
\label{tab:conf_tire}
\begin{tabular}{ccc}
\hline
\multicolumn{3}{c}{ \textbf{Test Set Confusion Matrix}}  \\
\hline
 & \textbf{Carpet}& \textbf{Cement}  \\
\hline
\textbf{Carpet} & \textbf{82.7 \%} & 17.3 \% \\
\textbf{Cement} & 14.8\% & \textbf{85.2 \%}\\
\hline
\end{tabular}
\end{center}
\end{minipage}
\begin{minipage}{0.45\linewidth}
\caption{ESP32 results online.}
\begin{center}
\label{tab:online_conf_tire}
\begin{tabular}{ccc}
\hline
\multicolumn{3}{c}{ \textbf{Online Confusion Matrix}}  \\
\hline
& \textbf{Carpet}& \textbf{Cement}  \\
\hline
\textbf{Carpet} & \textbf{81.8 \%} & 18.2 \% \\
\textbf{Cement} & 18.0 \% & \textbf{82.0\%}  \\
\hline
\end{tabular}
\end{center}
\end{minipage}
\label{tab:smth}
\end{table}

Table \ref{tab:online_conf_tire} shows the online results in the classification performed using a fully connected neural network with two ReLU hidden layers and an output layer that uses the softmax activation function. The overall classification rate is 81.9\%. An improvement in the neural network model is expected to result in an improvement in the online classification performance. In this online experiment we recorded 159 data windows in the carpet and 278 data windows in the cement. 


\subsubsection{Robotic Coatings}
\label{sec:skins}

Fig. \ref{fig:applicationfig}(c) shows a foam structure with embedded piezoelectric strip sensors and a top coat of fiberglass that sandwiches the sensor nodes. In this usage case, we aim to estimate a distance $\hat{r}$ and an angle $\hat{\theta}$ that represents how far away the source of an impact is, taking the center of the sensor node as the origin. The foam semi-sphere structure that is coated has an outer diameter of $20.32 cm$. 

 The neural network trained in Section \ref{sec:perf}, Test 1 was used in this problem. It consists of a convolution layer, followed by a maxpooling layer and two dense layers before the output. To perform this regression, we encode $r$ as a distance in cm and $\theta$ as the pair $(cos \theta, sin \theta)$ to achieve better performance through normalized outputs. In a preliminary test we obtain an online neural network accuracy of $1.49 \pm 1.5 cm$ at the center of the sensor node for a neural network that performed at testing set accuracy of  $3.8 cm$ (Euclidean distance). This accuracy decreases as we move further away from the center of the sensor node. When we use a dedicated neural network for each estimated state ($\{\hat{r}, \hat{\theta}\}$), we achieve a regression online error of $2.264\pm 1.384 cm$ for $\hat{r}$ and $0.758 \pm 0.543 rad$ for $\hat{\theta}$. Future work includes performing further tests in this platform to eliminate biases in the data collection and training and producing more robust estimations.

\subsubsection{Sensor calibration}
\label{sec:finger}

For this we use a combined proximity, contact and force (PCF) sensor that consists of an infrared proximity and barometer sensor embedded in an elastomer (rubber) layer developed by \cite{doi:10.1177/1687814019844643}. As the manufacturing process of the PCF sensor is not controlled, there are slight variations in the sensor readings (min and max values) after the polymer (Dragon Skin) is cured onto the bare ICs. Moreover, because of the dome shape of the sensor surface the infrared and barometer readings vary based on the position, orientation and magnitude of the contact. The viscoelastic nature of the rubber layer makes it challenging to analytically model the sensor behavior. Therefore, to estimate a single function with a fixed number of parameters that can map the raw barometer and infrared sensor readings to a true calibrated force is not a trivial task. Even though the Gaussian Processes Regression (GPR) that was used in the previous work \cite{doi:10.1177/1687814019844643} for calibration is the most accurate regression method, it has an exceptionally high computational complexity which prevents its usage for large numbers of samples or learning online. We therefore switch to \textit{nn4mc} to develop an embedded neural network to learn this mapping.


\begin{figure}[H]
\centering
\includegraphics[width=3cm, height=3cm, trim={1.7cm 0cm 1cm 10cm},clip, angle=0, keepaspectratio]{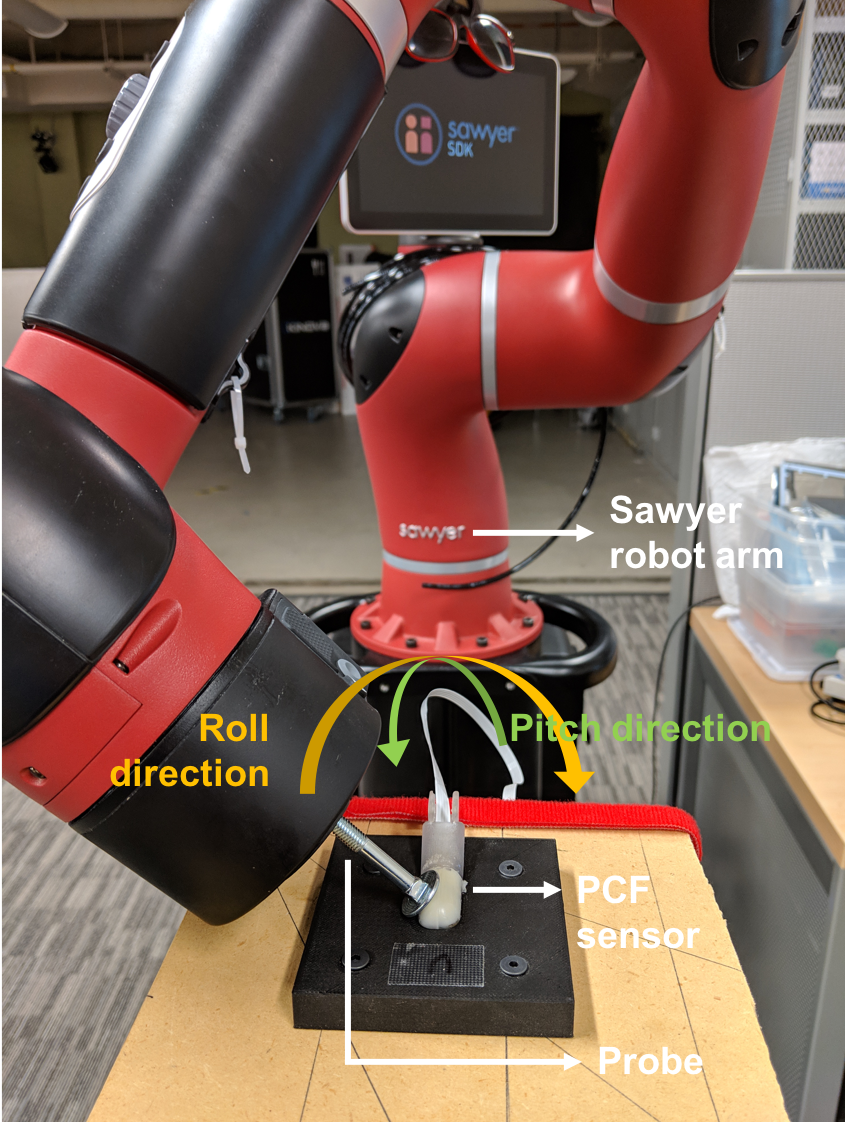}
\includegraphics[width=4.5cm, height=3.5cm, trim={2.5cm 0cm 3.5cm 0cm},clip, angle=0, keepaspectratio]{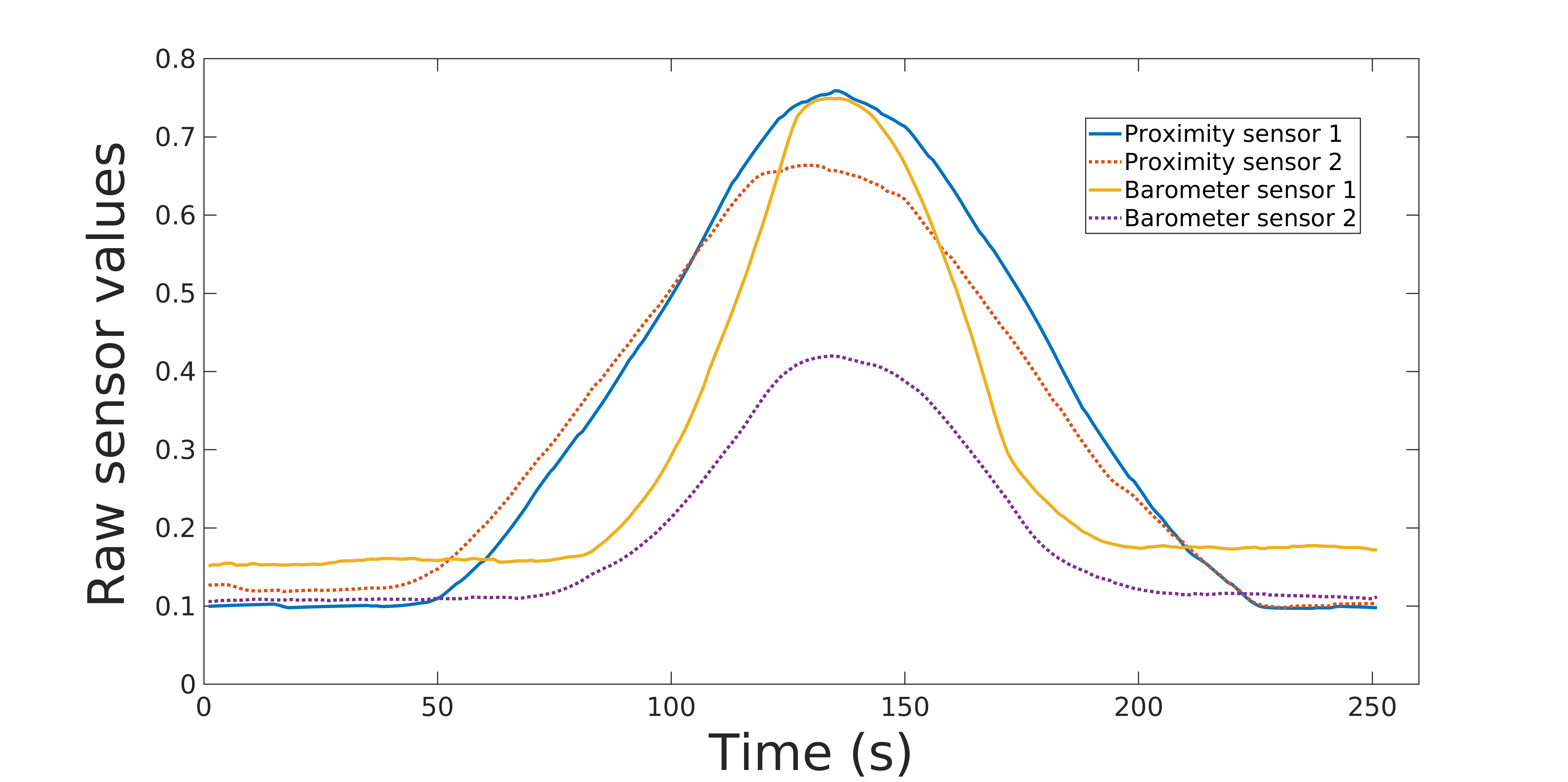}
\includegraphics[width=4.5cm, height=3.5cm, trim={2.5cm 0cm 3.5cm 0cm},clip, angle=0, keepaspectratio]{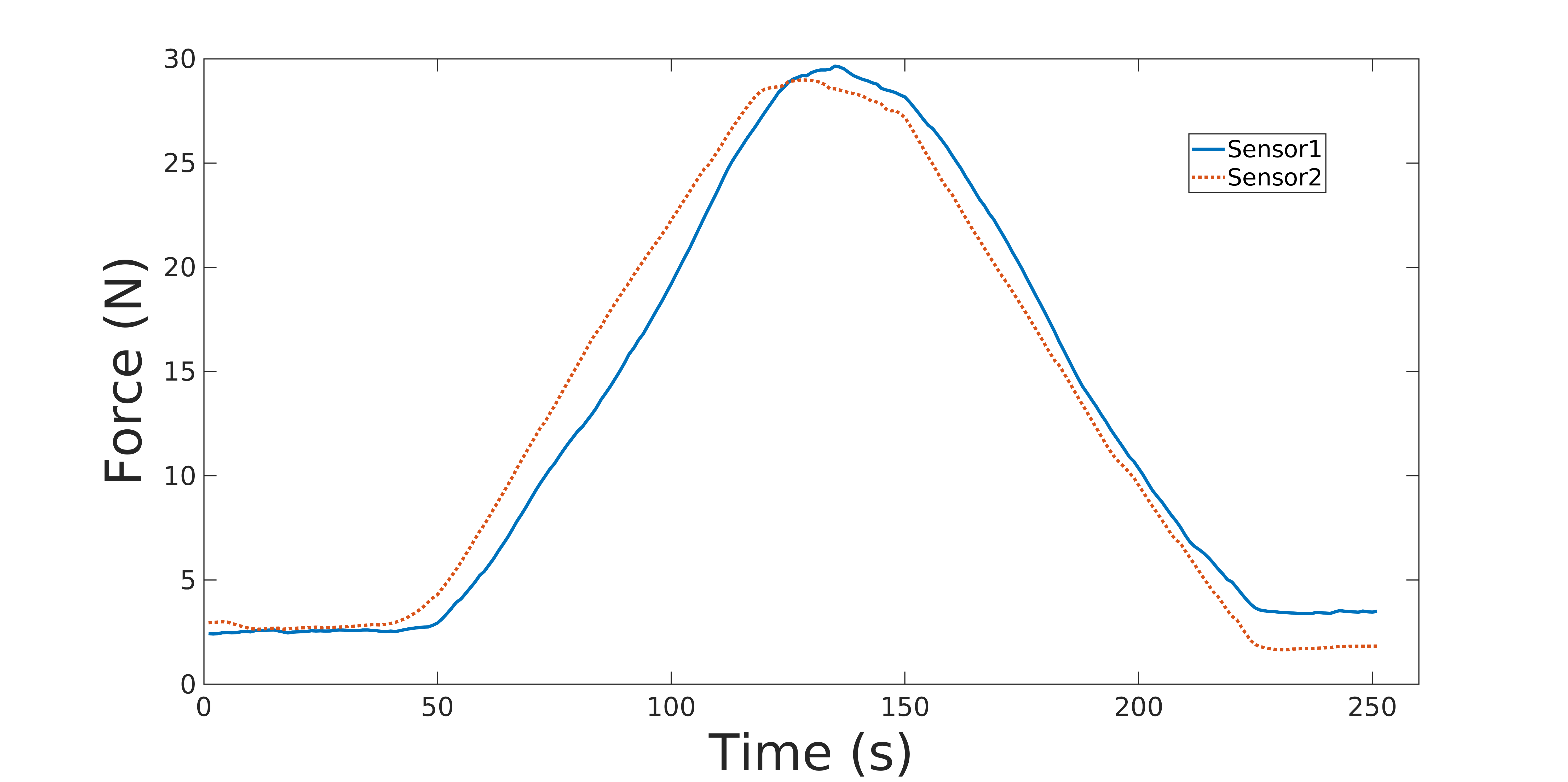}
\caption{\textbf{ \textit{Left:} Experimental setup for force calibration using the Sawyer
robot arm. \textit{Center:} Raw signals (proximity and barometer) from two PCF sensors. \textit{Right:} Calibrated force output from the neural network}}
\label{fig:exp_finger}
\end{figure}

For calibrating the PCF sensor, we automate the data collection process using the Sawyer robot arm (Figure \ref{fig:exp_finger} \textit{left}). A screw with a circular head is fixed on the end-effector of the Sawyer robot which acted like a probe. We program the Sawyer robot in impedance control mode to poke the sensor surface at discrete locations using the probe. The probe exerts increasing and decreasing force profiles onto the sensor surface from different roll and pitch angles (Figure  \ref{fig:exp_finger}). A 6-axis force torque sensor in the Sawyer's wrist acts as ground truth force measurements. The neural network is a simple feed forward, fully connected, network consisting of 5 layers. The input layer consists of 2 neurons, the second, third and fourth layer consists of 6, 12 and 4 neurons respectively and the last layer consists of 1 neuron. We used a rectified linear unit (ReLU) activation for first four layers and linear activation for the last layer.

The time to pass a single data point consisting of two raw sensor inputs to get a single force prediction takes around $2.00 \pm 0.06 ms$. The lightweight nature of the \textit{nn4mc} library allows us to run the neural network in real-time on an ATMega type (Arduino UNO) microprocessor for on-board calibration capability. More importantly, it allows us to have normalized and similar force profiles across multiple PCF sensors (Figure \ref{fig:exp_finger} \textit{center} and \textit{right}). The data shown in Figure \ref{fig:exp_finger} (\textit{center} and \textit{right}) is collected by manually applying (near to similar) forces on two of the robot fingers by pinching it between the thumb and index fingers with both of the experimenter's hands simultaneously. The raw data from the sensor and output from the neural network is parallely saved and plotted using MATLAB. Note that we have smoothed the data using the \textit{smoothdata} function in MATLAB for clarity purpose.  

\subsubsection{Neural Network Predictive Control Applications}
\label{sec:mpc}

In many applications we do not only want to make neural network predictions for signal processing; instead, we would like to make predictions on decisions that a control system should make based on its current input. This is true for cases where a plant model requires a forward numerical solution to a partial differential equation, which is more computationally expensive than forward passing data through a neural network. Another useful case is when we do not have a precise model of a plant's transfer function, but we have historical data on control input and plant output. 

\subsubsection{System Identification with Neural Networks}

 We simulate how a neural network would be used to learn the behavior of a nonlinear plant in order to make predictions on future steps based on the plant predictions on previous steps. To make a simple example, we simulate a plant that behaves as $y_p(x_t, u_t)= sin (x_t)$ and we generate uniformly random states $x(t) = \mathcal{U}(-2\pi; 2\pi)$ of varying duration synthetically.
 
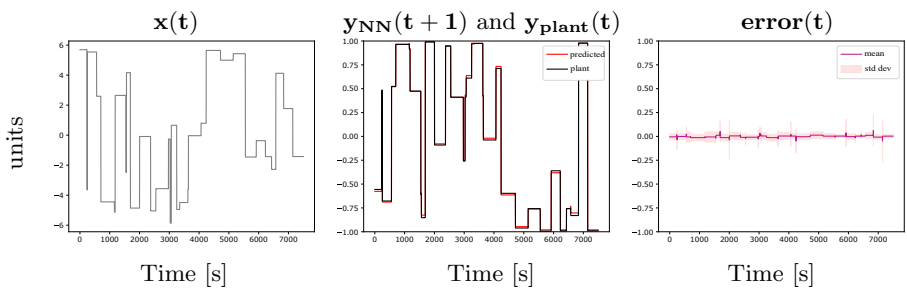
\begin{figure}
\centering
\includegraphics[width=\textwidth,trim={4cm 0cm 4cm 0cm},clip]{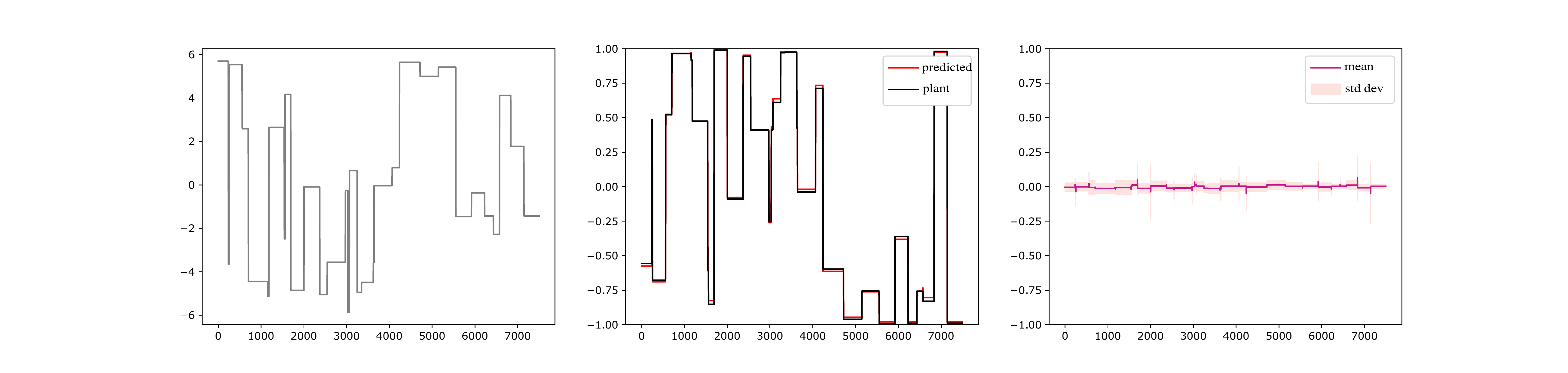}
 \begin{picture}(100,0)
      \put(-125, 47){\rotatebox{90}{units}}
          \put(-70,100){$\mathbf{x(t)}$}
          \put(1,100){$\mathbf{y_{NN}(t+1)}$ and $\mathbf{y_{plant}(t)}$}
           \put(150,100){$\mathbf{error(t)}$}
           \put(-75,5){Time [s]}
           \put(40,5){Time [s]}
            \put(145,5){Time [s]}
       \end{picture}
\caption{Results on simple neural network predictive system identification learned from former plant states. From left to right: randomly generated set of input states for testing the neural network; sample neural network prediction; mean errors and their standard deviation for 100 neural networks trained using the same architecture and different initialization.}
\label{fig:sys_id}
\end{figure}

In order to perform the neural network prediction, we create a shallow neural network with two hidden layers of 5 neurons each that takes as input the previous plant output and the current plant input to estimate what the future plant output should be. The error is then computed by taking the difference between the future time output and the prediction from the neural network. Given that the case study presented is artificial we compare the absolute error for the same neural network being trained 100 times with different initialization in Figure \ref{fig:sys_id}. 

\section{Discussion}

High performance microcontrollers with clock speeds in excess of 100 MHz are capable of performing forward prediction steps of complex CNNs in less than 10ms. With the ESP32 available as a system-on-a-module with a footprint of 18x25.5x3mm$^3$ with integrated wireless networking at a cost of less than four dollars, off-loading and distributing computation throughout a robotic system becomes feasible. Albeit equipped with a real-time platform, we observe higher variance in computational time than with a ``plain'' microcontroller such as the ARM Cortex M4 based Teensy 3.6 platform, albeit the observed variance might be negligible for most applications . 

Using embedded neural networks for classification is tightly coupled to embedded training \cite{EmbodiedNeuralNetwork}. With training smaller networks taking only a few minutes to train on a state-of-the-art desktop computer, embodied learning using back-propagation is potentially feasible on the platforms investigated here, albeit the memory requirements for learning ---in particular for recurrent neural networks that require backpropagation through time --- will quickly become prohibitive. Rather, robotic materials might take advantage of more powerful centralized hardware to collect data and improve the accuracy of embedded computation during down times of the robot. 

The proposed framework currently lacks the ability to balance accuracy and hardware limitations such as memory or computation time, both of which are directly related to the total number of layers and weights. Here, a growing body of work in  optimization of network architecture by pruning and even reorganization \cite{hughes2018material}, might not only allow to deal with existing constraints, but also inform the development of hardware computing architectures that are particularly suited to estimation and control problems that are specific to robotics.

Although \textit{nn4mc} provides a large class of microcontrollers with the ability to perform online neural network predictions, we are aware that the multiple-source-multiple-target nature of this package makes the estimation performance dependent on the selection of microcontroller from the user standpoint. One of the possible disadvantages is that the floating point math capabilities is dependent on the microcontroller. Therefore, there might exist a loss of accuracy near the decision boundary of a classification model that changes the classification decision completely. Another possible disadvantage is that floating point accuracy might be affected by the choice of activation function or post-processing functions, (i.e in Section \ref{sec:skins}). The primary disadvantage of this method is that some microcontrollers cannot support parameters for very deep models due to memory limitations, such as the Arduino UNO platform on Test 1. Finally, the abstract representation for the neural network opens the possibility for hardware layout generation.


\section{Conclusion and Future Work}
We presented an automated approach to generate signal processing and controls code for embedded microcontrollers using state-of-the-art machine learning tools. Based on our experience of hand-coding such algorithms (using both model-based and neural networks) for microcontrollers for the past 15 years, we find that this approach can dramatically reduce the development cycle for embedded signal processing applications and help increase research in the area of online low-level predictions. 

With regard to using such an approach to create a new class of smart composites, or ``robotic'' materials, we show that standard, low-cost microcontrollers that are capable of at least 16 bits fixed point arithmetic can achieve similar accuracy than desktop computers within time intervals that are acceptable for real-time computation, that is in the order of a few milliseconds, for a number of robotic applications.   

There is a trade-off between computation time, neural network architecture complexity and model fidelity. For example, by reducing the number of fixed point bits, we might get an improved neural network complexity allowance, but this might cause an adverse effect in the coarseness of the training parameters that might compromise the neural network accuracy. 

In future work, we wish to investigate the effects of weight pruning to further speed up computation and reduce the required memory, a recommendation engine for neural network architectures based on \cite{hughes2018material}, arbitrating centralized computational resources that can be used for learning among multiple embedded platforms, as well as include new sources and targets. 
We also expect to increase documentation on this software through tutorials, issue tracking on GitHub and a website. 

\subsubsection{Acknowledgements}

This research has been supported by the Air Force Office of Scientific Research (AFOSR), we are grateful for this support. We also thank Christoffer Heckman for allowing us to use his ninja car.

\bibliographystyle{ieeetr}
\bibliography{root} 

\end{document}